\documentclass[runningheads]{llncs}

\usepackage{ECCV}

\usepackage{eccvabbrv}

\usepackage{graphicx}
\usepackage{booktabs}

\usepackage[accsupp]{axessibility}  

\usepackage[pagebackref,breaklinks,colorlinks,citecolor=eccvblue]{hyperref}
\usepackage{hyperref}

\usepackage{orcidlink}

\usepackage{amsmath,amssymb}
\usepackage{algorithm}
\usepackage{algorithmic}
\usepackage{multirow}
\usepackage{threeparttable}
\usepackage{wrapfig}    

\begin{document}

\title{Clue Matters: Leveraging Latent Visual Clues to Empower Video Reasoning} 


\author{Kaixin Zhang\inst{1,2} \and
Xiaohe Li\inst{1} \and
Jiahao Li\inst{1,2} \and
Haohua Wu\inst{1,2} \and
Xinyu Zhao\inst{1,2} \and
Zide Fan\inst{1,2} \and
Lei Wang\inst{1,2}
}
\authorrunning{Zhang et al.}

\institute{Aerospace Information Research Institute, Chinese Academy of Sciences\and
University of Chinese Academy of Sciences}

\maketitle

\begin{abstract}
Multi-modal Large Language Models (MLLMs) have significantly advanced video reasoning, yet Video Question Answering (VideoQA) remains challenging due to its demand for temporal causal reasoning and evidence-grounded answer generation. Prevailing end-to-end MLLM frameworks lack explicit structured reasoning between visual perception and answer derivation, causing severe hallucinations and poor interpretability. Existing methods also fail to address three core gaps: faithful visual clue extraction, utility-aware clue filtering, and end-to-end clue-answer alignment. Inspired by hierarchical human visual cognition, we propose ClueNet, a clue-aware video reasoning framework with a two-stage supervised fine-tuning paradigm without extensive base model modifications. Decoupled supervision aligns clue extraction and chain-based reasoning, while inference supervision with an adaptive clue filter refines high-order reasoning, alongside lightweight modules for efficient inference. Experiments on NExT-QA, STAR, and MVBench show that ClueNet outperforms state-of-the-art methods by $\ge 1.1\%$, with superior generalization, hallucination mitigation, inference efficiency, and cross-backbone compatibility. This work bridges the perception-to-generation gap in MLLM video understanding, providing an interpretable, faithful reasoning paradigm for high-stakes VideoQA applications.
\end{abstract}

\section{Introduction}
Recent advances in Multi-modal Large Language Models (MLLMs) have driven remarkable breakthroughs in video reasoning for a wide range of high-stakes applications, including embodied robotics, intelligent surveillance, and human-computer interaction~\cite{zhang2025videollama, wang2025internvideo2, bai_Qwen25VL_2025}. A core and uniquely challenging task is Video Question Answering (VideoQA), which requires processing temporally continuous content, comprehending evolving entities and causal interrelations across frames, and generating accurate, evidence-grounded answers~\cite{xiao2021next, wu2024star, li2024mvbench, lei2018tvqa}.

Despite impressive performance on short-form benchmarks, prevailing MLLM-based frameworks follow a monolithic end-to-end paradigm: aligning visual tokens with textual representations via feature projectors~\cite{cheng2024videollama, lin2024video, zhang2023video} or cross-modal interaction modules~\cite{lin2024video, wang2024internvideo2}, and generating answers in a one-shot ``intuitive'' manner. This omits an explicit, structured reasoning step bridging low-level perception and high-level answer derivation, leading to two fundamental limitations: (1) dense temporal visual content exacerbates catastrophic hallucinations~\cite{wang2024videohallucer, li2023evaluating, guan2024hallusionbench}; and (2) the black-box pipeline suffers from severely limited interpretability. For example, when answering ``Which object did the person pick up after setting down the mug?'', state-of-the-art (SOTA) models frequently hallucinate by ignoring the temporal order clue while exposing no reasoning process for validation.
\begin{figure}[t]
  \centering
  \includegraphics[width=0.92\linewidth]{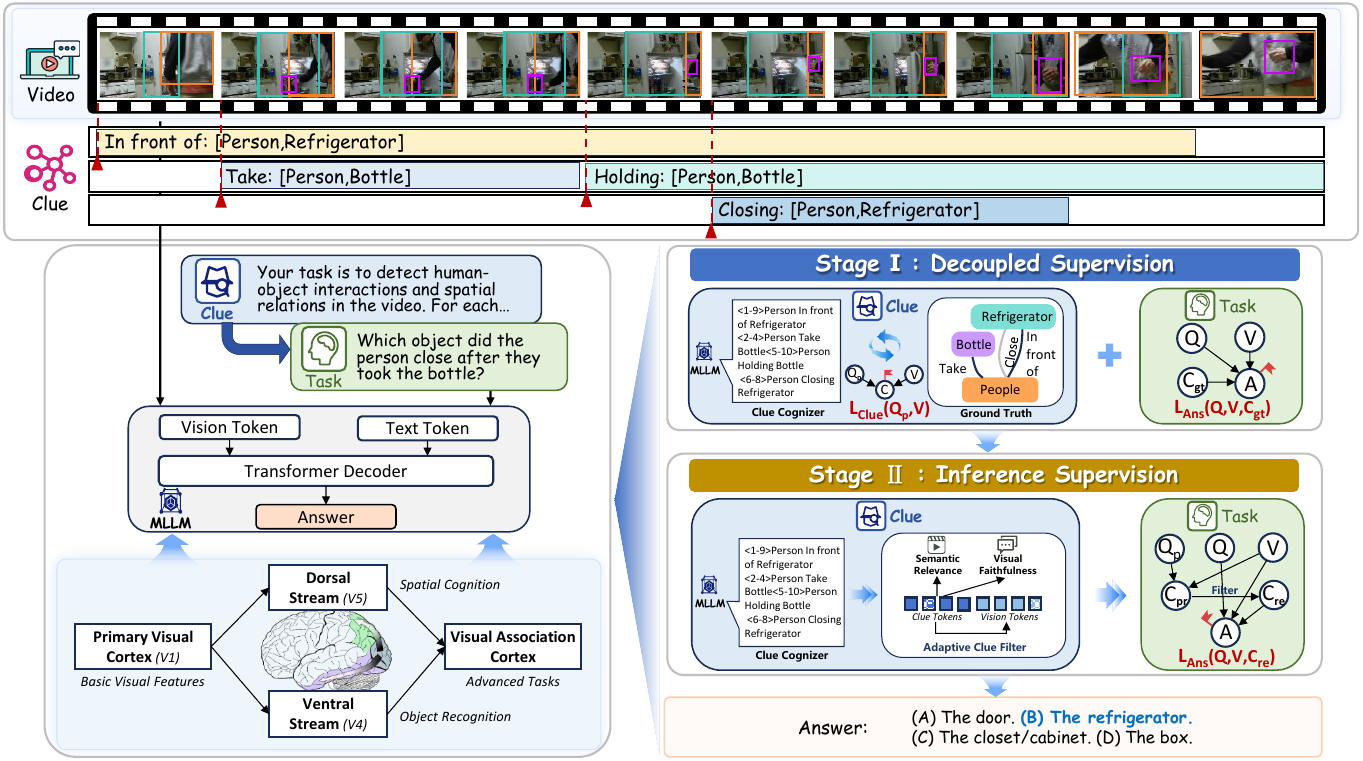}
  \caption{Overall framework of the proposed ClueNet, which mimics the hierarchical human visual cognitive process to enable explicit clue-aware video reasoning.}
  \label{fig:framework}
  \vspace{-1em}
\end{figure}
Recent works attempt to enhance VideoQA reasoning: CCoT~\cite{mitra2024compositional} constructs frame-level scene graphs via CoT~\cite{wei2022chain} prompting; VoT~\cite{fei2024video} introduces a scene graph module for step-by-step reasoning; VideoEspresso~\cite{han2025videoespresso} provides a 200K-sample CoT fine-tuning dataset. Yet none fundamentally addresses faithful visual clue extraction and utility-aware clue judgment. Existing approaches overlook three critical gaps: (1) lack of faithful, accurate temporal visual clue mining causes missed critical event dynamics and spurious temporal reasoning; (2) missing explicit supervision for clue relevance and utility assessment leads to reasoning path error propagation; and (3) disjoint clue extraction and answer generation lack end-to-end alignment between intermediate clues and final outputs. We conduct a detailed analysis of the causes of reasoning failures in existing MLLMs in Section~3.

Inspired by the hierarchical mechanism of human visual cognition~\cite{felleman1991distributed, dicarlo2012does}, where V1 handles low-level parsing, V2--V5 extract perceptual clues~\cite{mishkin1983object}, and visual association areas coordinate with the prefrontal cortex to filter and integrate clues for high-order reasoning~\cite{miller2001integrative, buschman2007top}, we propose \textbf{ClueNet}: a reliable video reasoning framework that discovers, filters, and leverages latent visual clues via a two-stage supervised fine-tuning paradigm, with no extensive base model modifications.

In the first stage, \textit{Decoupled Supervision} jointly optimizes semantic clue extraction and clue chain-based reasoning. Via instruction tuning~\cite{liu2023visual}, ClueNet builds a full-video dynamic clue cognizer to track cross-temporal entity states, ensuring end-to-end alignment between clue extraction and downstream reasoning. In the second stage, \textit{Inference Supervision} fine-tunes high-order reasoning to ground answers strictly in structured clue chains. We iteratively refine reasoning with model-mined raw video clues, and propose an Adaptive Clue Filter to enhance critical clues via a differentiable gating mechanism guided by Semantic Relevance and Visual Faithfulness. The refined clue chain is iteratively input to the model, enabling explicit step-by-step reasoning without the computational overhead of RL-based selection. 

When inferencing, Keyframe Selection and Visual Compression modules implement preliminary keyframe extraction and clue-aligned token selection respectively, ensuring inference efficiency and strengthened clue-visual consistency.

Extensive experiments on NExT-QA~\cite{xiao2021next}, STAR~\cite{wu2024star}, and MVBench~\cite{li2024mvbench} demonstrate that our proposed ClueNet outperforms SOTA video reasoning methods by at least 1.1\% across all benchmarks , with strong test-time generalization on MVBench, consistently robust hallucination mitigation, significantly accelerated inference efficiency, and universal generalizability of its core clue-mining reasoning paradigm across diverse mainstream MLLM backbones. Our core contributions are summarized as:
\begin{itemize}
    \item We propose \textbf{ClueNet}, the first framework to unify hierarchical clue extraction supervision and utility-aware filtering into an end-to-end training-inference pipeline for reliable MLLM video reasoning.
    \item We design a progressive two-stage fine-tuning paradigm with Decoupled Supervision for clue-reasoning alignment and Inference Supervision for high-order reasoning enhancement, at minimal training cost.
	\item We develop an Adaptive Clue Filter to suppress spurious clue-induced hallucinations, alongside an inference-time optimization pipeline comprising training-free Keyframe Selection and clue-aligned Visual Compression for improved inference efficiency and clue-visual consistency.
	\item Extensive experiments on NExT-QA, STAR, and MVBench show ClueNet 
    outperforms SOTA by $\ge1.1\%$, with strong generalization, hallucination 
    mitigation, accelerated inference, and cross-backbone generalizability.
\end{itemize}

\section{Related Work}

\textbf{MLLMs for Video Reasoning}
VideoQA demands spatio-temporal reasoning to align dynamic visual sequences with linguistic queries. Early works use recurrent neural networks or 3D CNNs~\cite{lei2018tvqa, jang2017tgif}, advancing to spatio-temporal attention and graph-based memory networks~\cite{le2020hierarchical, xiao2021next}. The field has since shifted to MLLMs~\cite{zhu2023minigpt, liu2023visual}, with video-specific models (Video-ChatGPT~\cite{maaz2024video}, Video-LLaVA~\cite{lin2024video}, VideoLLaMA series~\cite{zhang2023video, cheng2024videollama, zhang2025videollama}) mapping sampled frames to LLM embedding space via visual encoders and alignment modules. Despite strong generalization, these monolithic architectures treat spatio-temporal reasoning as a black box, causing severe temporal hallucinations~\cite{li2023evaluating, guan2024hallusionbench, li2025vidhalluc, wang2024videohallucer} driven by language priors rather than grounded visual evidence.

\noindent\textbf{Efficient Video Context Reduction}
Unconstrained video processing incurs prohibitive computational cost due to extreme spatio-temporal redundancy, addressed mainly via frame sampling or token compression. Frame-level methods ~\cite{zhao2025efficient, yu2024frame, ren2024timechat} leverage motion priors, text-conditioned querying or temporal grounding to outperform naive uniform sampling~\cite{maaz2024video}, but rely on disjointed multi-stage pipelines or rigid external components, adding overhead and breaking end-to-end differentiability. Token-level pruning strategies ~\cite{zhang2025beyond, huang2025prunevid, yuan2025dtoma, chen2024image} dynamically remove redundant visual features during encoding, but their visual-centric design lacks structured scene abstraction, risking irreversible loss of prompt-critical fine-grained details. Existing paradigms lack a unified end-to-end mechanism to preserve high-value visual evidence, leaving critical semantics vulnerable to premature, unrecoverable elimination.

\noindent\textbf{Intermediate Reasoning and Symbolic Representations}
Structured intermediate representations bridge low-level perception and cognitive reasoning. Entity relationship or scene graph, a formalism for encoding objects and their relations, is established as a core video understanding task via Action Genome~\cite{ji2020action}. Subsequent works advance relational modeling with hierarchical graph architectures~\cite{nguyen2024hig}, object-centric relational encoders~\cite{herzig2018mapping}, and multi-step graph reasoning for VideoQA~\cite{mao2022dynamic,jiang2020reasoning}, but these methods decouple graph construction from downstream reasoning, limiting joint optimization. For structured reasoning without explicit graph supervision, Multimodal Chain-of-Thought approaches use scene-graph-guided prompting~\cite{mitra2024compositional}, sequential reasoning decomposition~\cite{fei2024video}, and multi-step visual rationale generation~\cite{zhang2023multimodal}, yet their reasoning chains lack robust anchoring to verifiable spatio-temporal locations, making them error-prone on spatially demanding tasks.

\section{Question Analysis}
\label{sec:analysis}
This section systematically dissects the failure causes of canonical VideoQA models and the core limitations of existing MLLMs. All diagnostic visualizations and case studies are validated on mainstream open-source VideoLLaMA3~\cite{zhang2025videollama} and Qwen2.5-VL~\cite{bai_Qwen25VL_2025} for generality. We categorize failures into three types of cognitive biases: visual perception bias, clue cognition bias, and inductive reasoning bias.
\begin{figure}[t]
  \centering
  \includegraphics[width=0.9\linewidth]{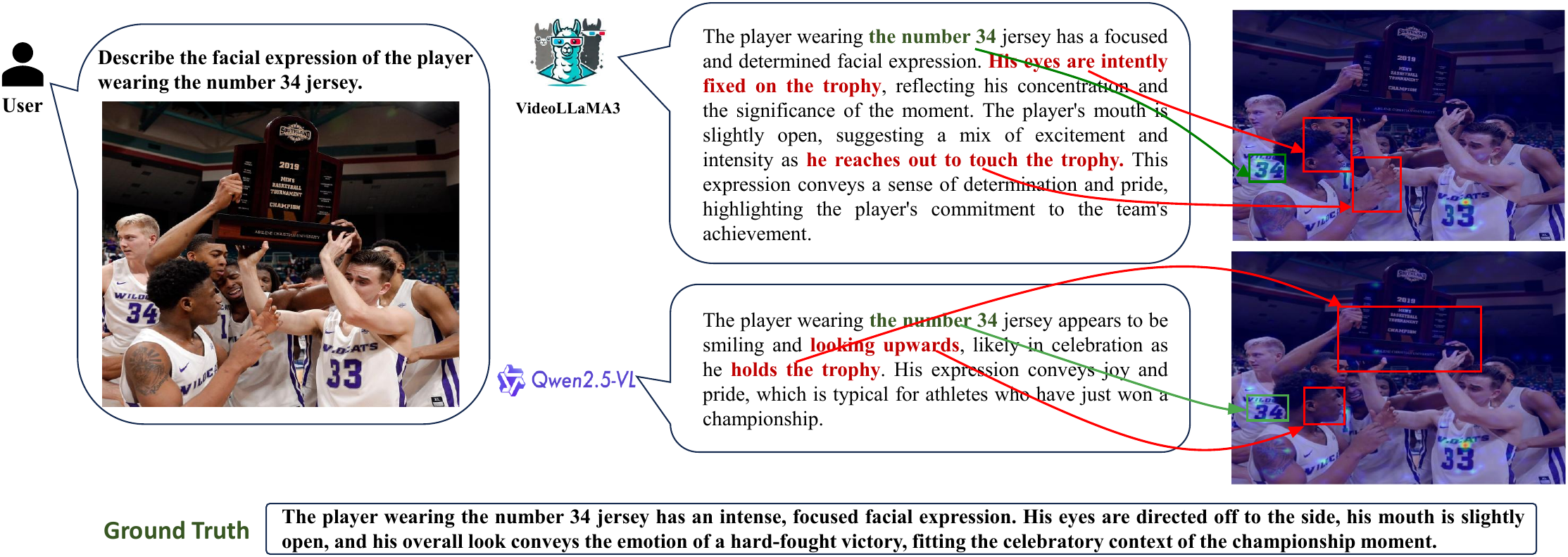}
  \caption{MLLM attention heatmaps and paired VideoQA results: misaligned attention strongly correlates with incorrect answers, confirming visual perception bias.}
  \label{fig:question1}
  \vspace{-1em}
\end{figure}
\paragraph{Visual Perception Bias.}

Human primary visual cortex uses task-driven selective attention to rapidly identify query-relevant high-salience visual clues and adaptively allocate attention to critical regions, even for sparse, transient cross-frame clues~\cite{desimone1995neural}. Existing video understanding methods fail to 
systematically emulate this query-guided, temporally-aware attention paradigm. We validate this via attention heatmap visualization (Fig.~\ref{fig:question1}), plotting layer-averaged cross-attention weights of VideoLLaMA3 and Qwen2.5-VL. In the visualized case, the model misassociates a critical query clue (jersey number) with the wrong subject, producing erroneous evidence that triggers cascaded reasoning errors. This limitation motivates us to reorient the model’s input visual tokens toward valid, query-critical regions and strengthen coupling between fine-grained visual features and high-level logical reasoning clues.
\begin{figure}[t]
  \centering
  \includegraphics[width=0.95\linewidth]{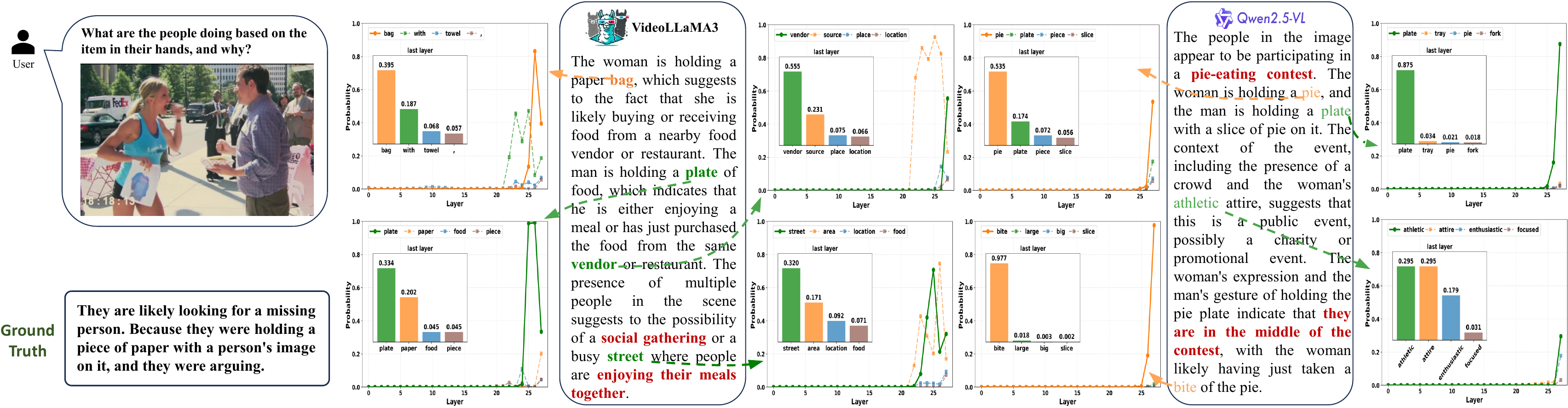}
  \caption{Intermediate layers top-probability output token visualization: erroneous visual clue semantic interpretation in intermediate layers propagates to hallucinated reasoning and incorrect final answers, confirming clue cognition bias.}
  \label{fig:question2}
  \vspace{-1em}
\end{figure}
\paragraph{Clue Cognition Bias.}
Human higher-order visual cortex transforms raw perceptual inputs into linguistically structured, semantically coherent interpretations, and uses working memory to maintain cross-frame temporal consistency of visual episodes for evidence-aligned descriptions~\cite{milner2008two}. Existing MLLMs lack systematic modeling of this clue-to-semantic mapping, 
resulting in a ``knowing \emph{what} but not \emph{why}'' phenomenon: superficially plausible answers without grounded, interpretable logical support. As shown in Fig.~\ref{fig:question2}, VideoLLaMA3 and Qwen2.5-VL misclassify a woman’s held portrait poster as a bag or pie, triggering fully erroneous reasoning. Without explicit clue semantic grounding, these misinterpretations propagate through the generation pipeline to cause final VideoQA errors, motivating our proposed clue cognizer module that maps accurately perceived visual clues to verifiable semantic interpretations.
\begin{figure}[t]
  \centering
  \includegraphics[width=0.8\linewidth]{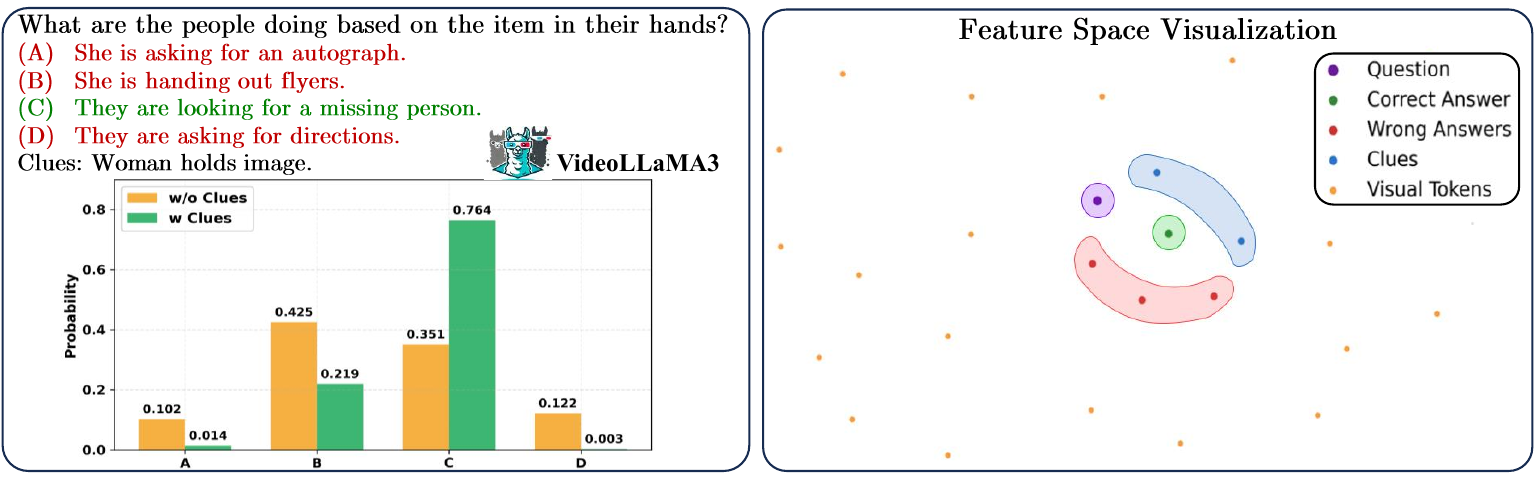}
  \caption{t-SNE visualization of question, correct/incorrect candidate answer, reasoning clues, and sampled visual token features: clue-answer feature proximity, causal reasoning utility, and ground-truth clue performance boost confirm inductive reasoning bias.}
  \label{fig:question3}
    \vspace{-1em}
\end{figure}
\paragraph{Inductive Reasoning Bias.}
The human prefrontal cortex and visual association regions integrate cross-frame visual evidence with reasoning chains, generate evidence-aligned answers, and self-verify consistency across perception, reasoning, and conclusions~\cite{miller2001integrative}. Existing MLLMs rarely emulate this hierarchical cross-modal fusion paradigm, leading to weak consistency constraints between clue representations, reasoning chains, and predictions. This decouples perception, logic, and final outputs, drastically elevating hallucination risk. We validate this via feature space analysis (Fig.~\ref{fig:question3}) and comparative experiments: query-relevant valid clues have significantly higher similarity to ground-truth answers than spurious clues, and guiding models with these valid clues consistently improves reasoning faithfulness and accuracy. These findings confirm that reliable video reasoning demands faithful evidence grounding, motivating our approach to leverage query-aligned clues.

\section{Method}
\subsection{Framework Overview}
The ClueNet architecture overview is shown in Fig.~\ref{fig:framework}, with two core components: (1) a Clue Cognizer, which uses instruction prompting to guide an MLLM to generate structured, temporally grounded latent visual clue descriptions; (2) a principled two-stage training scheme that maximizes extracted clues' utility for video understanding, while enforcing cognitive consistency between visual perception and high-level reasoning.
\subsection{Clue Cognizer}
The Clue Cognizer transforms raw visual inputs into structured logical evidence via a dynamic $\mathcal{G}$ for the full video, where each visual clue $C_i \in \mathcal{G}$ is a structured quintuplet:
\begin{equation}
    \label{eq:clue}
    C_i = \langle t_{\text{start}},\, t_{\text{end}},\, s_i,\, r_i,\, o_i \rangle,
\end{equation}
with $s_i$, $o_i$ as subject and object entities, $r_i$ the interaction predicate, and $[t_{\text{start}}, t_{\text{end}}]$ the temporal frame lifespan. Unlike static image scene graphs~\cite{krishna2017visual, herzig2018mapping}, this representation explicitly tracks the temporal evolution of entity interactions, optimized for temporal VideoQA reasoning~\cite{ji2020action, nguyen2024hig}. For instance, a generated clue is instantiated as: \texttt{"<2-4> person take bottle"}. A complete clue example is provided in Fig.~\ref{fig:framework}. To ensure temporal consistency, the collection is updated incrementally across video segments to maintain alignment with the evolving scene state.

During training, we supervise clue generation with a standard autoregressive language modeling objective using timestamp-aligned scene annotation tuples. Let $c_{i} = f_{\theta}(Q_P,V)$ denote the $i$-th token of the generated clue sequence, where $V$ is the input video and $Q_P$ the full-video clue extraction instruction prompt. The visual clue generation loss for an $L$-frame video is:
\begin{equation}
	\mathcal{L}_{\mathrm{clue}} = -\frac{1}{L}\sum_{i=1}^{L}\ell_{CE}\left(C_{gt,i}, f_{\theta}(Q_P, V)\right),
\end{equation}
where $\ell_{CE}$ is cross-entropy loss. We minimize $\mathcal{L}_{\mathrm{clue}}$ to produce video- and question-grounded intermediate reasoning descriptions, with the resulting multi-frame clue sequence fed into the downstream answer generation module.
\begin{figure}[t]
  \centering
  \includegraphics[width=0.9\linewidth]{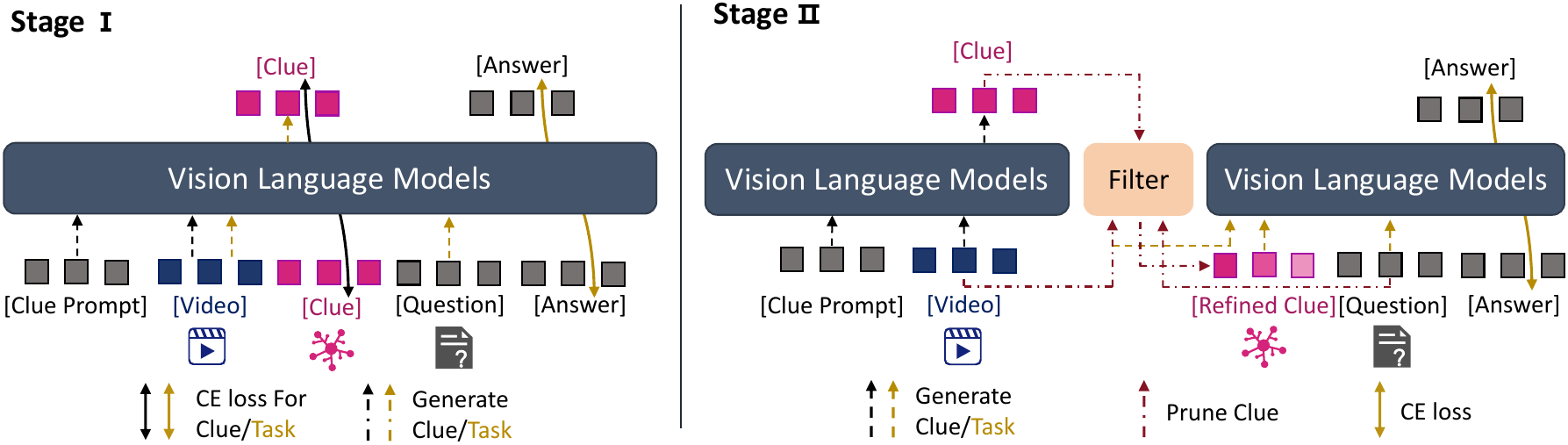}
  \caption{Overview of our two-stage training scheme.}
  \label{fig:stage}
  \vspace{-1em}
\end{figure}

\subsection{Stage 1: Decoupled Supervision of Clue Discovery and Reasoning}
To enable reliable endogenous clue discovery and utilization, we decouple video reasoning into two complementary sub-tasks: (1) video-based clue discovery, and (2) clue-grounded answer generation, optimized with separate dedicated supervision signals.

To enforce exclusive clue-grounded reasoning, the model is trained to generate answers conditioned on the question, ground-truth clues, and video content. The Stage 1 training objective is:
\begin{equation}
	\mathcal{L}_{\mathrm{Stage1}} = \mathcal{L}_{\mathrm{clue}} + \ell_{CE}\left(A, f_{\theta}(C_{gt}^{full}, Q, V)\right),
\end{equation}
where $C_{gt}^{full}$ is the full ground-truth clue token sequence. The two loss terms correspond to the decoupled sub-tasks: the first supervises accurate video-based clue generation, the second supervises ground-truth clue-guided answer generation. To stabilize training and avoid conflicting gradients, we optimize the two terms alternately per batch.
\subsection{Stage 2: Adaptive Clue Filter and Inference Supervision}
Stage 1 decoupled supervision trains effective ground-truth clue generation and utilization, but does not resolve the critical training-inference gap: during end-to-end inference, the model relies on self-generated clues, which may contain noise, irrelevant content, or hallucinations from imperfect visual grounding and MLLM backbone knowledge gaps. To close this gap and enforce end-to-end perception-reasoning consistency, we introduce the Adaptive Clue Filter (ACF), which retains only high-fidelity, question-relevant clues.

\begin{figure}[t]
  \centering
  \includegraphics[width=0.9\linewidth]{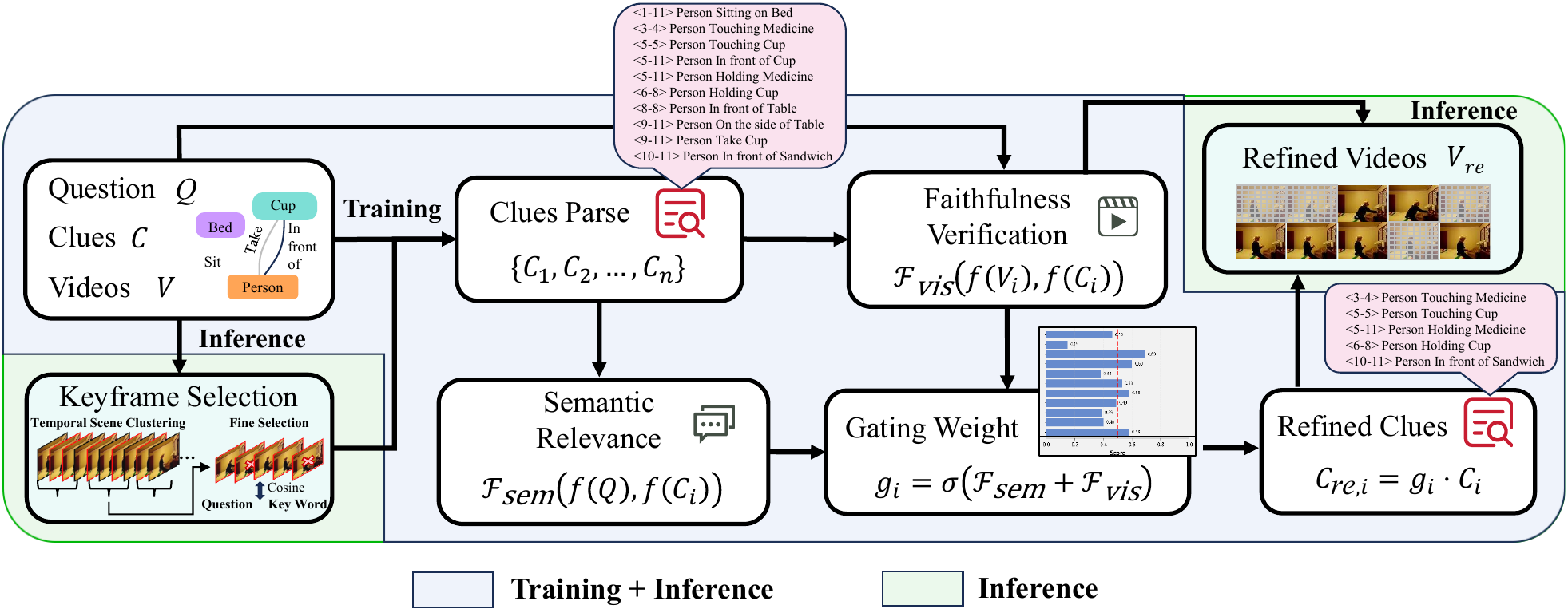}
  \caption{Refining clues and visual information during the training and inference stages.}
  \label{fig:filter}
  \vspace{-1.5em}
\end{figure}

After obtaining candidate scene clues $C_{pr}$ via the Clue Cognizer, ACF evaluates each candidate $C_i$ along two dimensions to ensure question relevance and visual grounding. Since each clue $C_i$ is grounded to a specific temporal segment, we use $V_i$ to denote the sequence of frames encompassed by $C_i$.
\begin{itemize}
    \item \textbf{Semantic Relevance}: Question informativeness, measured by alignment between the MLLM final-layer average token representations of the clue $f(C_i)$ and question $f(Q)$.
    \item \textbf{Visual Faithfulness}: Visual evidence support, measured by alignment between $f(C_i)$ and $f(V_i)$, where $f(V_i)$ is the average visual token representation of frames in $V_i$ from the visual encoder.
\end{itemize}
The final clue gating weight $g_i$ is computed via a two-branch network (Fig.~\ref{fig:filter}):
\begin{equation}
	g_i = \sigma \left(\mathcal{F}_{\mathrm{sem}}\!\left(f(Q), f(C_i)\right) + \mathcal{F}_{\mathrm{vis}}\!\left(f(V_i), f(C_i)\right) \right),
\end{equation}
where $\mathcal{F}_{\mathrm{sem}}$ and $\mathcal{F}_{\mathrm{vis}}$ each consist of linear layer, LayerNorm, and ReLU activation, and $\sigma(\cdot)$ is the sigmoid function. Learned gating weights scale the input embeddings of all tokens within each variable-length clue, yielding refined representations $C_{\mathrm{re}} = \{g_i\cdot C_i\}$. We add $\ell_1$ sparsity regularization on gating weights to encourage focus on critical, non-redundant evidence. The Stage 2 training objective is:
\begin{equation}
	\mathcal{L}_{\mathrm{Stage2}} = \ell_{CE}\left(A, f_{\theta}(C_{re}^{full}, Q, V)\right) + \frac{\lambda}{N} \sum_{j=1}^{N} \left|g_j\right|,
\end{equation}
where $\lambda$ is the loss balance hyperparameter, and $C_{re}^{full}$ is the full refined clue token sequence. This end-to-end supervision makes $g_i$ a differentiable filter enforcing logical consistency from clue generation to answer induction, closing the training-inference gap and maintaining clue-driven reasoning with self-generated inference-time clues.

\subsection{Inference Process}
The complete inference pipeline sequentially performs Keyframe Selection, 
Clue Generation, ACF, and Visual Compression before final answer generation, 
as illustrated in Fig.~\ref{fig:filter}.

\noindent \textbf{Keyframe Selection (KS).}
To reduce spatio-temporal redundancy in long videos and sample critical keyframes, we propose a two-step Keyframe Selection module:
\begin{itemize}
    \item \textbf{Redundancy elimination}: We adopt LVNet’s Temporal Scene Clustering (TSC) algorithm~\cite{park2024too}, extracting frame-level features via pre-trained ResNet-18~\cite{he2016deep}, then grouping visually similar frames via TSC’s dynamic thresholding to produce a condensed candidate set $V_{c}$. We follow the original hyperparameter settings.
    \item \textbf{Semantic-driven Fine Selection}: For each frame in $V_{c}$, we extract patch-level visual embeddings $E_v \in \mathbb{R}^{N \times P \times d_{v}}$ ($N$ = candidate frames, $P$ = patches per frame, $d_v$ = embedding dimension). We use spaCy~\cite{honnibal2020spacy} to select nouns/verbs as semantic keywords from the question and obtain their initial text token set as $E_k$ (assuming $|E_k|=J$), plus the global question embedding $E_q$ from the MLLM’s textual encoder. The relevance score $S_i$ for frame $i\in[1,N]$ is computed via patch-level max-pooling:
\begin{equation}
S_i = \alpha\max_{1\le p\le P}\text{sim}(E_q, E_{v}^{(i, p)}) + \beta \max_{1\le j\le J,1\le p\le P} \text{sim}(E_{k}^{(j)}, E_{v}^{(i, p)}).
\end{equation}
where $\text{sim}(\cdot, \cdot)$ denotes cosine similarity, and $\alpha, \beta$ balance global question and local keyword alignment. We select the top 16 frames by $S_i$ to form the Keyframe $V'$.
\end{itemize}

\noindent\textbf{Visual Compression (VC).}
Given a cropped video $V'$ and a question $Q$, the MLLM generates initial clues $C_{pr}$. which are refined by ACF to get refined clues $C_{re}$. Formally, we reuse the Visual Faithfulness score $s_{v}^{(i,t)}$ from $\mathcal{F}_{\mathrm{vis}}$, which measures the alignment between clue $C_i$ and frame $V_t \in V'$. Since multiple clues may overlap temporally, the retention score for $V_t$ is defined by averaging the products of gating weights and Visual Faithfulness scores across all clues encompassing it:
\begin{equation}
S_{\text{ret}}(V_t) = \frac{1}{|\mathcal{M}_t|} \sum_{i \in \mathcal{M}_t} g_i \cdot s_{v}^{(i,t)},
\end{equation}
where $\mathcal{M}_t = \{ i \mid C_i \in C_{pr},\ t \in [t_{i,\text{start}}, t_{i,\text{end}}] \}$. We retain frames exceeding a unified threshold $\tau$:
\begin{equation}
V_{re} = \{ V_t \in V^{'} \mid S_{\text{ret}}(V_t) \ge \tau \}.
\end{equation}

The LLM ingests interleaved $C_{re}$ and $V_{re}$ to generate the final answer.

\subsection{Theoretical Analysis}
We provide an information-theoretic justification for \textsc{ClueNet} via the Information Bottleneck (IB) principle~\cite{tishby2000information}. Let $V$, $Q$, $C$, and $A$ denote the input video, question, extracted clues, and target answer, respectively. We formulate $C$ as a task-relevant bottleneck variable that mediates the reasoning process. The IB principle seeks a minimal sufficient representation by maximizing predictive power $I(C;A)$ while minimizing input redundancy $I(C; V, Q)$, yielding the objective:
\begin{equation}
    \mathcal{L}_{\mathrm{IB}}(C) = I(C; V, Q) - \gamma I(C; A).
    \label{eq:theoretical_ib}
\end{equation}
where $\gamma > 0$ balances input compression and predictive power.

Fundamentally, all core components of ClueNet are designed to optimize this trade-off: our decoupled Stage 1 training maximizes the sufficiency term $I(C; A)$, while the ACF and evidence-driven visual compression aggressively minimize the redundancy term $I(C; V, Q)$. 

\section{Experiments}
\subsection{Experimental Setup}
\textbf{Datasets and Benchmarks}
We evaluate ClueNet on three challenging video reasoning benchmarks:
\begin{itemize}
    \item \textbf{NExT-QA}~\cite{xiao2021next}: VideoQA benchmark with Multi-Choice (MC) and Open-Ended (OE) tasks across Causal, Temporal, and Descriptive question types.
    \item \textbf{STAR}~\cite{wu2024star}: Real-world situated reasoning benchmark with four splits (\textit{Interaction}, \textit{Sequence}, \textit{Prediction}, \textit{Feasibility}), requiring evidence-grounded logical reasoning.
    \item \textbf{MVBench}~\cite{li2024mvbench}: Multi-modal evaluation suite with 4,000 QA pairs across 20 spatio-temporal tasks, spanning low-level visual perception to high-level cognitive reasoning.
\end{itemize}

For both NExT-QA and STAR, scene graph clue annotations are provided 
and converted into our structured quintuplet format (Eq.~\ref{eq:clue}).

\textbf{Metrics.} For multiple-choice tasks (NExT-QA MC, STAR, MVBench), we report standard \textbf{Accuracy (\%)}. For the open-ended NExT-OE task, we adopt Wu-Palmer Similarity (WUPS)~\cite{malinowski2014multi}, which measures semantic similarity between predictions and ground truths based on taxonomic depth. We report both the lenient WUPS@0 (retaining raw scores for any semantic overlap) and the strict WUPS@0.9 (0.9 similarity threshold, penalizing imprecise predictions with 0.1× scaling).

\noindent\textbf{Baselines}
We compare against state-of-the-art models, spanning both closed-source models (GPT-4o, Qwen-VL-Max) and open-source video-language models, Results for Flipped-VQA~\cite{ko2023large}, ViLA~\cite{wang2024vila}, and VideoChat2~\cite{li2024mvbench} are from their original papers. All other open-source baselines, Video-LLaVA~\cite{lin2024video}, VideoLLaMA3~\cite{zhang2025videollama}, 
LongVA~\cite{zhang2024long}, Qwen2.5-VL~\cite{bai_Qwen25VL_2025}, 
Qwen3-VL~\cite{bai2025qwen3}, Cos~\cite{hu2025cos}, 
InternVideo2.5~\cite{wang2025internvideo2} are re-implemented and evaluated with a maximum of 16 input frames to ensure fair comparison.

\noindent\textbf{Implementation Details}
We adopt VideoLLaMA3~\cite{zhang2025videollama} as our base model for fine-tuning. The visual encoder is a frozen pre-trained SigLIP-SO400M~\cite{zhai2023sigmoid}; the LLM is initialized with Qwen2.5-7B-Instruct ~\cite{qwen_Qwen25_2025}. For fair comparison, VideoLLaMA3 is also fine-tuned for one round of standard supervised fine-tuning (SFT) (input: $V$, $Q$; output: $A$) under identical training settings.

\subsection{Main Results}
\begin{table}[!t]
    \centering
    \caption{SOTA method comparison on STAR and NExT-QA: multiple-choice task accuracy ($\%$). Best and second-best results are \textbf{bolded} and \underline{underlined}. Total (Tot) denotes the overall accuracy over all test samples.}
    \label{tab:star_nextqa}
    \footnotesize
    \resizebox{0.63\linewidth}{!}{
    \begin{tabular}{lccccccccc}
        \toprule
        \multirow{2}{*}{Model} & \multicolumn{5}{c}{STAR} & \multicolumn{4}{c}{NExT-QA} \\
        \cmidrule(lr){2-6} \cmidrule(lr){7-10}
        & Int & Seq & Pred & Feas & Tot & Cau & Tem & Des & Tot \\
        \midrule
        \multicolumn{10}{c}{\textit{Closed-source MLLMs}} \\
        \cmidrule(lr){1-10}
        GPT-4o        & 66.5 & 75.3 & 70.2 & 67.1 & 71.1 & 80.1 & 73.8 & 84.7 & 78.9 \\
        Qwen-VL-Max   & 67.6 & 74.5 & 67.3 & 65.8 & 70.7 & 79.7 & 74.1 & 84.3 & 78.7 \\
        \midrule
        \multicolumn{10}{c}{\textit{Open-source MLLMs}} \\
        \cmidrule(lr){1-10}
        LongVA~\cite{zhang2024long}        & 50.2 & 54.5 & 57.2 & 54.3 & 53.3 & 56.9 & 62.8 & 65.8 & 60.2 \\
        Video-LLaVA~\cite{lin2024video}   & 40.0 & 41.1 & 45.7 & 42.9 & 41.3 & 55.2 & 53.6 & 69.4 & 57.0 \\
        Flipped-VQA~\cite{ko2023large}   & 66.2 & 67.9 & 57.2 & 52.7 & 65.4 & 72.7 & 69.2 & 75.8 & 72.0 \\
        ViLA~\cite{wang2024vila}          & \underline{70.0} & 70.4 & 65.9 & 62.2 & 69.2 & 75.3 & 71.8 & 82.1 & 75.3 \\
        VideoChat2~\cite{li2024mvbench}   & 58.4 & 60.9 & 55.3 & 53.1 & 59.0 & 57.4 & 61.9 & 69.9 & 60.8 \\
        Cos~\cite{hu2025cos}           & 50.7 & 57.3 & 62.0 & 54.6 & 55.2 & 67.3 & 66.1 & 69.0 & 67.2 \\
        InternVideo2.5~\cite{wang2025internvideo2} & 63.8 & 73.6 & 67.2 & \textbf{68.0} & 69.1 & 81.0 & 76.5 & 87.0 & 80.6 \\
        Qwen2.5-VL~\cite{bai_Qwen25VL_2025}    & 63.5 & 73.3 & 65.9 & 64.2 & 68.4 & 81.5 & \underline{79.0} & 86.5 & \underline{81.5} \\
        Qwen3-VL ~\cite{bai2025qwen3}     & 54.4 & 59.1 & 60.3 & 58.2 & 57.5 & 77.1 & 71.5 & 83.1 & 76.3 \\
        VideoLLaMA3~\cite{zhang2025videollama}   & 68.6 & \underline{75.6} & \underline{71.0} & 67.7 & \underline{72.1} & \underline{81.7} & 77.2 & \underline{87.8} & 81.3 \\
        \textbf{ClueNet (Ours)} & \textbf{75.8} & \textbf{80.7} & \textbf{77.6} & \underline{66.7} & \textbf{77.6} & \textbf{83.5} & \textbf{83.7} & \textbf{88.1} & \textbf{84.3} \\
        \bottomrule
    \end{tabular}
    }
\vspace{-1em}
\end{table}

\subsubsection{Quantitative Evaluation on STAR and NExT-QA}
As shown in Table~\ref{tab:star_nextqa}, ClueNet achieves state-of-the-art (SOTA) performance on both benchmarks: 77.6\% on STAR and 84.3\% on NExT-QA, outperforming all open-source and closed-source competitors. It gains +5.5\% over VideoLLaMA3 (STAR) and +2.8\% over Qwen2.5-VL (NExT-QA), validating the superiority of explicit clue-based reasoning over pure model scaling.

The gains are most pronounced where temporal and causal understanding is required. On \textit{Sequence} and \textit{Temporal} splits, ClueNet outperforms VideoLLaMA3 by +5.1\% and Qwen2.5-VL by +4.7\%, confirming that two-stage training directs the model to reason over temporal clue trajectories rather than static visual shortcuts. On \textit{Interaction} and \textit{Causal} splits, the dynamic clue collection in the Clue Cognizer disambiguates complex entity-action relationships that confound end-to-end baselines (+5.8\% over ViLA, +1.8\% over VideoLLaMA3). The sole exception is the \textit{Feasibility} split: ClueNet trails InternVideo2.5 by 1.3\%, as its questions demand unobservable commonsense knowledge, creating a minor, acceptable trade-off from the ACF’s evidence-first filtering given gains elsewhere.
\begin{table}[!t]
    \centering
    \caption{Evaluation on the NExT-OE generative task using WUPS. Best and second-best results are \textbf{bolded} and \underline{underlined}.}
    \label{tab:next_oe}
    \resizebox{0.9\linewidth}{!}{
\begin{tabular}{lcccccccc}
        \toprule
        \multirow{3}{*}{Model} & \multicolumn{8}{c}{NExT-OE} \\
        \cmidrule{2-9}
        & \multicolumn{2}{c}{All} & \multicolumn{2}{c}{Causal} & \multicolumn{2}{c}{Temporal} & \multicolumn{2}{c}{Descriptive} \\
        \cmidrule(lr){2-3} \cmidrule(lr){4-5} \cmidrule(lr){6-7} \cmidrule(lr){8-9}
        & WUPS@0 & WUPS@0.9 & WUPS@0 & WUPS@0.9 & WUPS@0 & WUPS@0.9 & WUPS@0 & WUPS@0.9 \\
        \midrule
        LongVA         & 27.7 & 21.3 & 20.8 & 14.2 & 22.3 & 16.1 & 50.3 & 44.1 \\
        Cos            & 27.8 & 21.4 & 21.0 & 14.3 & 22.3 & 16.3 & 50.4 & 44.3 \\
        Video-LLaVA    & 25.6 & 19.5 & 17.9 & 12.0 & 21.7 & 15.5 & 47.9 & 41.5 \\
        InternVideo2.5 & 28.4 & 22.0 & 25.1 & 18.1 & \underline{27.3} & 20.1 & 37.4 & 33.1 \\
        Qwen2.5-VL     & 30.6 & 24.6 & 23.2 & 17.1 & 25.0 & 18.6 & \underline{54.6} & \underline{49.6} \\ 
        Qwen3-VL       & 29.1 & 23.1 & 22.0 & 15.8 & 23.7 & 17.3 & 52.2 & 46.7 \\ 
        VideoLLaMA3    & \underline{32.3} & \underline{25.7} & \underline{25.6} & \underline{18.4} & \underline{27.3} & \underline{20.2} & 53.9 & 48.7 \\
        \textbf{ClueNet(Ours)}  & \textbf{35.5} & \textbf{29.2} & \textbf{26.2} & \textbf{19.3} & \textbf{28.9} & \textbf{22.2} & \textbf{64.9} & \textbf{60.5} \\
        \bottomrule
    \end{tabular}
    }
\vspace{-1em}
\end{table}
\vspace{-1em}
\subsubsection{Open-Ended Generation Performance}
Table~\ref{tab:next_oe} reports results on the NExT-OE generative task. Unlike multi-choice tasks, open-ended generation demands precise free-form visual grounding. To comprehensively evaluate semantic accuracy, we report both WUPS@0 and WUPS@0.9. ClueNet consistently achieves state-of-the-art performance across all categories under both metrics. The most striking gain is in \textit{Descriptive}: ClueNet reaches WUPS@0/0.9 of 64.9/60.5, outperforming Qwen2.5-VL by a significant margin ($+10.3$/$+10.9$). This improvement stems from ClueNet’s ability to guide the LLM to focus on fine-grained object attributes and state transitions.
ClueNet also demonstrates remarkable resilience to the strict $\alpha=0.9$ penalty on \textit{Temporal} and \textit{Causal} tasks, confirming that structured clue chains provide a reliable symbolic anchor that mitigates hallucination in long-form event reasoning.
\subsubsection{Generalization Evaluation on MVBench}
To verify the generalization performance of ClueNet in the out-of-distribution scenario, we conduct experiments on the MVBench benchmark (Table~\ref{tab:mvbench}), which comprises 20 fine-grained video understanding tasks. ClueNet achieves state-of-the-art performance with an average score of 69.8\%, outperforming strong open-source baselines such as InternVideo2.5 (68.7\%) and VideoLLaMA3 (68.3\%), while also surpassing closed-source models like GPT-4o. 

ClueNet leads on dynamic tasks where frame-averaging baselines typically struggle: Action Prediction (AP: 79.5\%) and Object Interaction (OI: 83.0\%) are both top-ranked, confirming that temporally-grounded clues faithfully track entity state transitions across frames. Strong performance on counting-centric tasks (AC: 57.5\%, MC: 78.0\%) further demonstrates that the instance-level dynamic visual clues preserve discriminative temporal structure in complex scenes. In static attribute tasks such as Object Existence, where temporal reasoning offers little advantage, ClueNet is competitive but does not lead, consistent with our design focus on temporal dynamics.
\begin{table}[t]
    \centering
    \caption{Evaluation on the MVBench benchmark. Best and second-best results are \textbf{bolded} and \underline{underlined}, respectively. ClueNet achieves the highest average score, excelling particularly in fine-grained temporal and object interaction tasks.}
    \label{tab:mvbench}
    \resizebox{0.98\textwidth}{!}{
    \begin{tabular}{lrrrrrrrrrrrrrrrrrrrrr}
        \toprule
        Model          & Avg   & AS    & AP    & AA    & FA    & UA    & OE    & OI    & OS    & MD    & AL    & ST    & AC    & MC    & MA    & SC    & FP    & CO    & EN    & ER    & CI    \\
        \midrule
        \multicolumn{22}{c}{\textit{Closed-source MLLMs}} \\
        \midrule
        GPT-4o         & 60.8  & 65.5  & 75.5  & 82.5  & 48.5  & \underline{83.5} & 69.5  & 72.0  & 37.5  & 46.0  & 41.5  & 90.0  & 49.0  & 46.0  & 70.5  & 57.5  & 47.5  & 72.0  & 41.0  & 59.5  & 61.0  \\
        Qwen-VL-Max    & 61.4  & 64.5  & 77.0  & 82.0  & \underline{50.0} & \textbf{85.5} & 74.0  & 75.5  & 36.0  & 52.0  & 33.5  & 91.0  & 47.0  & 54.5  & 64.5  & 61.0  & 46.5  & 74.0  & \textbf{45.0} & \underline{62.5} & 51.5  \\
        \midrule
        \multicolumn{22}{c}{\textit{Open-source MLLMs}} \\
        \midrule
        Cos            & 49.6  & 57.5  & 58.0  & 57.0  & 44.0  & 69.0  & 50.5  & 61.0  & 32.0  & 37.5  & 35.0  & 86.5  & 48.5  & 27.5  & 58.0  & 47.5  & 44.5  & 58.5  & 39.5  & 57.0  & 42.0  \\
        LongVA         & 49.4  & 53.5  & 59.0  & 67.0  & 43.5  & 67.5  & 50.0  & 59.0  & 33.0  & 35.0  & 33.0  & 85.5  & 50.5  & 30.0  & 58.5  & 49.5  & 43.5  & 59.0  & 39.0  & 51.0  & 41.0  \\
        VideoChat2     & 51.1  & 66.0  & 47.5  & 83.5  & 49.5  & 60.0  & 58.0  & 71.5  & \underline{42.5} & 23.0  & 23.0  & 88.5  & 39.0  & 42.0  & 58.5  & 44.0  & 49.0  & 36.5  & 35.0  & 40.5  & 65.5  \\
        Video-LLaVA    & 42.3  & 44.0  & 52.0  & 54.5  & 39.0  & 53.5  & 45.0  & 45.5  & 38.5  & 27.0  & 24.5  & 84.5  & 33.0  & 27.0  & 51.0  & 41.0  & 42.0  & 51.5  & 31.5  & 42.5  & 37.5  \\
        InternVideo2.5 & \underline{68.7} & \textbf{83.5} & 70.0  & \textbf{91.0} & 48.0  & 75.5  & 89.5  & 77.5  & 41.0  & \underline{64.0} & 41.0  & 90.0  & \underline{57.0} & 77.0  & \textbf{95.0} & \underline{67.0} & \underline{51.0} & 72.0  & \underline{41.5} & \underline{62.5} & \textbf{80.0} \\
        Qwen2.5-VL     & 65.2  & \underline{76.5} & 70.5  & 84.5  & 48.0  & 79.0  & \textbf{94.0} & 71.0  & 34.5  & 53.0  & 42.0  & 91.5  & 41.5  & \textbf{81.0} & 91.5  & 55.0  & 50.0  & 77.5  & 39.0  & 53.0  & 71.5  \\
        Qwen3-VL       & 64.5  & 64.5  & 58.0  & 78.5  & 48.5  & 78.0  & 82.0  & 60.5  & 42.0  & \textbf{66.5} & 35.5  & 82.5  & 37.5  & 67.5  & 87.0  & 59.0  & 48.0  & 71.0  & 36.0  & 55.0  & 64.0  \\
        VideoLLaMA3    & 68.3  & 73.5  & \underline{77.5} & 83.0  & \underline{50.0} & \underline{83.5} & 92.5  & \underline{79.0} & 42.0  & 56.0  & \underline{48.0} & \underline{92.0} & 56.5  & \underline{78.5} & 91.5  & \textbf{67.5} & \underline{51.0} & \textbf{79.5} & 32.0  & 57.5  & 74.5  \\
        \textbf{ClueNet(Ours)}  & \textbf{69.8} & \underline{76.5} & \textbf{79.5} & \underline{90.0} & \textbf{51.5} & 80.5  & \underline{93.5} & \textbf{83.0} & \textbf{43.0} & 58.5  & \textbf{48.5} & \textbf{92.5} & \textbf{57.5} & 78.0  & \underline{92.0} & 65.5  & \textbf{51.5} & \underline{79.0} & 35.0  & \textbf{64.0} & \underline{77.0} \\
        \bottomrule
    \end{tabular}
    }
\vspace{-1em}
\end{table}
\subsection{Ablation Study}
Table~\ref{tab:ablation} reports ablations on STAR and NExT-QA using VideoLLaMA3~\cite{zhang2025videollama} as base model (Row 1).

\textbf{Two-Stage Training Paradigm.} The performance drop 
in Stage~1 alone ($-1.3\%$ on STAR) empirically illustrates the Inductive 
Reasoning Bias identified in Sec.~\ref{sec:analysis}: the model overfits 
to perfect semantic anchors and its reasoning chain collapses when 
inference-time clues introduce inevitable noise. Stage~2 inference 
supervision acts as a critical regularization mechanism, forcing the model 
to deduce answers robustly from imperfect, self-generated evidence, 
yielding a significant $+3.0\%$ recovery and gain on STAR.

\textbf{Keyframe Selection (KS) \& Adaptive Clue Filter (ACF).} As a heuristic pre-processing step, KS yields a modest $+0.3\%$ gain, 
suggesting its benefit is primarily realized in conjunction with downstream 
clue filtering rather than as a standalone component. Integrating 
the trainable ACF yields a substantial $+2.5\%$ surge on STAR by rigorously 
gating clues based on Visual Faithfulness and Semantic Relevance, directly 
mitigating the Clue Cognition Bias and preventing hallucinations from 
propagating into the final reasoning process.

\textbf{Visual Compression (VC).} 
The ${\sim}21\%$ reduction in GFLOPs ($51.4$K to $40.6$K) and acceleration to $3.93$s with only a negligible $-0.3\%$ accuracy trade-off empirically validates our Information Bottleneck hypothesis: once reasoning is anchored to ACF-verified clues, unanchored visual tokens become safely removable without loss of task-critical semantics.
\begin{table}[!t]
    \centering
    \caption{ClueNet ablation on STAR and NExT-QA; deltas relative to Row 1 (VideoLLaMA3~\cite{zhang2025videollama}) baseline. Row 1 inference time omitted (different pipeline).}
    \label{tab:ablation}
    {\footnotesize
    \resizebox{0.6\textwidth}{!}{
    \begin{tabular}{@{\extracolsep{\fill}} ccccc|cccc @{}}
        \toprule
        \multicolumn{5}{c|}{Components} & \multicolumn{2}{c}{Benchmarks} & \multicolumn{2}{c}{Inference} \\
        \cmidrule(lr){1-5} \cmidrule(lr){6-7} \cmidrule(lr){8-9} 
        Stage 1 & Stage 2 & KS & ACF & VC & STAR & NExT-QA & Time (s) & GFLOPs \\
        \midrule
        - & - & - & - & - & 72.1 & 81.3 & - & - \\
        \checkmark & - & - & - & - & 70.8 \textcolor{gray}{(-1.3)} & 80.6 \textcolor{gray}{(-0.7)} & 6.5060 & 54,197.96 \\
        \checkmark & \checkmark & - & - & - & 75.1 \textcolor{green}{(+3.0)} & 83.3 \textcolor{green}{(+2.0)} & 5.9963 & 54,026.85 \\
        \checkmark & \checkmark & \checkmark & - & - & 75.4 \textcolor{green}{(+3.3)} & 83.4 \textcolor{green}{(+2.1)} & 6.0443 & 54,075.02 \\
        \checkmark & \checkmark & \checkmark & \checkmark & - & \textbf{77.6} \textcolor{green}{(+5.5)} & \textbf{84.3} \textcolor{green}{(+3.0)} & 4.9963 & 51,355.89 \\
        \checkmark & \checkmark & \checkmark & \checkmark & \checkmark & 77.3 \textcolor{green}{(+5.2)} & 84.1 \textcolor{green}{(+2.8)} & \textbf{3.9298} & \textbf{40,622.84} \\
        \bottomrule
    \end{tabular}
    }
    \vspace{-1em}
    }
\end{table}
\subsection{In-Depth Analysis of Leveraging Clues}
To locate the primary bottleneck (high-level reasoning vs. low-level visual perception) of current VideoQA systems, we conduct an oracle clue experiment on STAR. Oracle experiments supply perfect ground-truth intermediate information to eliminate upstream errors and isolate downstream component performance; our setup feeds models ground-truth dynamic clue collections with video and question inputs, bypassing visual clue extraction to set a perfect perception upper bound. 

\begin{wrapfigure}{r}{0.48\textwidth}
  \centering
    \vspace{-15pt} 
  \includegraphics[width=\linewidth]{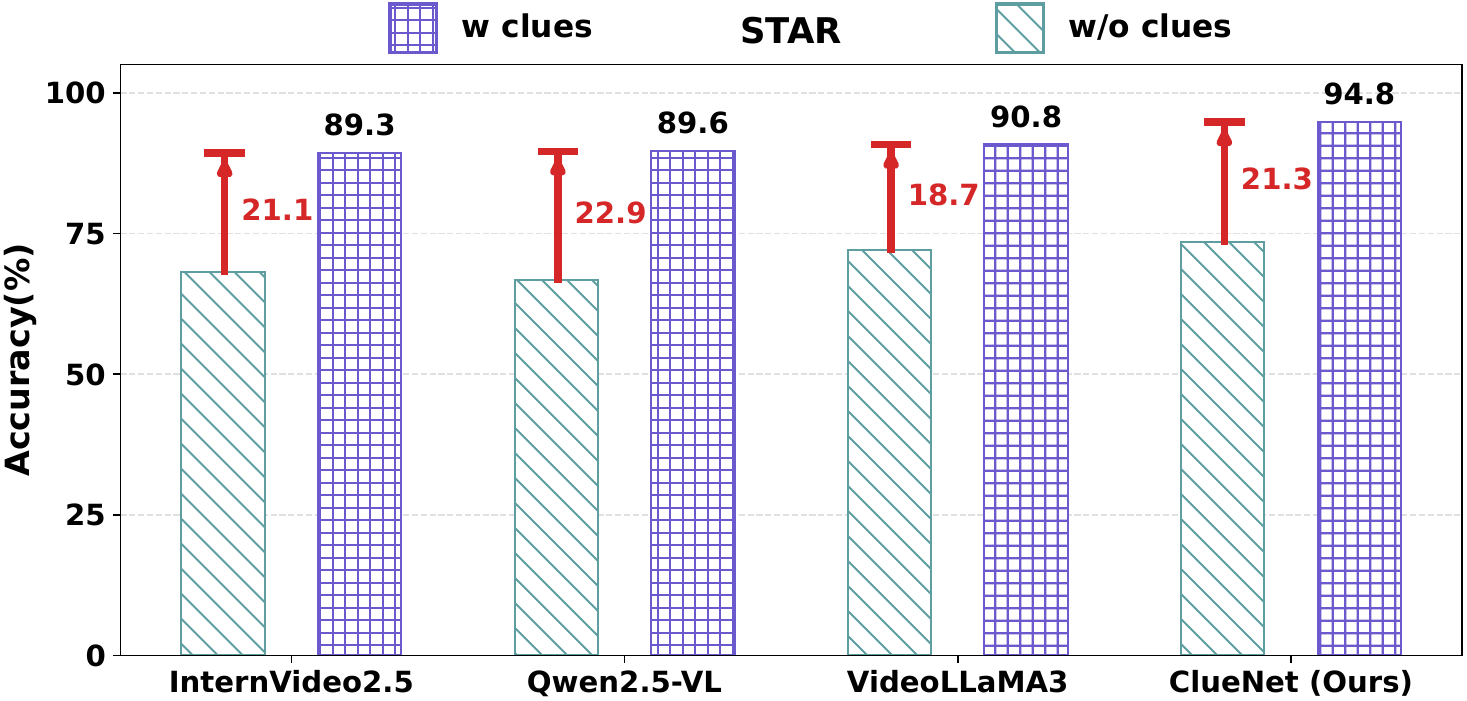} 
    \vspace{-15pt} 
  \caption{Oracle experiment results on STAR. Striped and gridded bars denote standard and oracle inference.}
  \label{fig:oracle_comparison}
\end{wrapfigure}


As shown in Fig.~\ref{fig:oracle_comparison}, all models exhibit dramatic accuracy improvements with oracle clues: Qwen2.5-VL and InternVideo2.5 improve by $+22.9\%$ and $+21.1\%$, respectively. The near-uniform gain magnitude (${\sim}20\%$) across architecturally diverse backbones confirms that these failures are systemic properties of the end-to-end paradigm itself, rather than artifacts of any specific architecture. This validates ClueNet's core design philosophy: explicitly mining structured logical evidence is a model-agnostic remedy for the cognitive biases identified, and the clue-aware pipeline constitutes a general reasoning paradigm for addressing the perception-cognition gap in current video understanding systems.

Notably, ClueNet achieves the highest oracle accuracy of $94.8\%$, 
outperforming all baselines by at least $+4.0\%$ despite a comparable delta 
gain (${+21.3\%}$), confirming that Stage 2 inference supervision 
optimizes the model's ability to exploit structured clues beyond clue quality 
alone. The residual ${\sim}5.2\%$ error gap under oracle conditions corresponds to the Feasibility split underperformance in Table~\ref{tab:star_nextqa}, where unobservable commonsense knowledge poses an irreducible reasoning limit orthogonal to visual clue quality.

\section{Conclusion}
We present ClueNet, a clue-aware video reasoning framework bridging visual perception and logical deduction. Decomposing QA into explicit visual clue localization and logical clue cognition, it mitigates hallucinations prevalent in standard end-to-end models. Extensive experiments on STAR, NExT-QA, and MVBench validate its state-of-the-art performance, with strong generalization, hallucination mitigation, accelerated inference, and cross-backbone generalizability.

\bibliographystyle{splncs04}
\bibliography{main}

\end{document}